\documentclass[letterpaper]{article} 
\usepackage{aaai24}  
\usepackage{times}  
\usepackage{helvet}  
\usepackage{courier}  
\usepackage[hyphens]{url}  
\usepackage{graphicx} 
\usepackage{natbib}  
\usepackage{caption} 
\usepackage{algorithm}
\usepackage{algorithmic}
\usepackage{newfloat}
\usepackage{listings}
\DeclareCaptionStyle{ruled}{labelfont=normalfont,labelsep=colon,strut=off} 
\floatstyle{ruled}
\newfloat{listing}{tb}{lst}{}
\floatname{listing}{Listing}
\usepackage{algorithm}
\usepackage{algorithmic}
\usepackage{amsmath}
\usepackage{amssymb}
\usepackage{booktabs}
\usepackage{multirow}

\title{SFC: Shared Feature Calibration in Weakly Supervised Semantic Segmentation}
\author{
    Xinqiao Zhao\textsuperscript{\rm 1,}\textsuperscript{\rm 2}\equalcontrib,
    Feilong Tang\textsuperscript{\rm 1}\equalcontrib,
    Xiaoyang Wang\textsuperscript{\rm 1,}\textsuperscript{\rm 2,}\textsuperscript{\rm 3},
    Jimin Xiao\textsuperscript{\rm 1}\thanks{Corresponding author.}
} 
\affiliations{
    \textsuperscript{\rm 1}Xi'an Jiaotong-Liverpool University\\
    \textsuperscript{\rm 2}University of Liverpool\\
    \textsuperscript{\rm 3}Metavisioncn\\

    xqz@liverpool.ac.uk,
    Feilong.Tang19@xjtlu.edu.cn,
    wangxy@liverpool.ac.uk,
    Jimin.Xiao@xjtlu.edu.cn
}

\usepackage{bibentry}

\begin{document}

\maketitle

\maketitle
\begin{figure*}[t!]
\centering
\includegraphics[height=5.7cm]{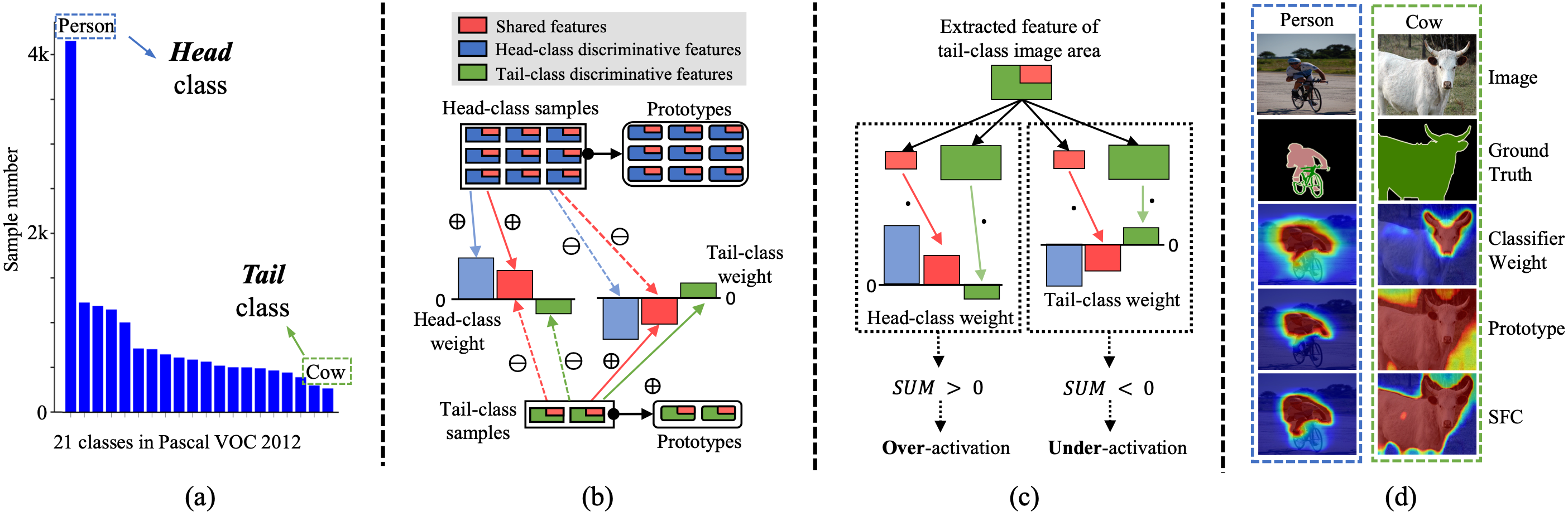}
\caption{Illustration of how shared features influence CAMs under a long-tailed scenario and the effects of our proposed SFC. (a) shows Pascal VOC 2012 \cite{everingham2010pascal} is a naturally long-tailed distributed dataset. (b) explains the shared feature components in head-/tail-class classifier weights and prototypes. (c) shows how over-/under-activations happen. (d) shows the CAMs of head-/tail-class examples.
Our SFC achieves better results with appropriate activation areas.} 
\label{intro}
\end{figure*} 
\begin{abstract}
Image-level weakly supervised semantic segmentation has received increasing attention due to its low annotation cost. Existing methods mainly rely on \textbf{C}lass \textbf{A}ctivation \textbf{M}apping (CAM) to obtain pseudo-labels for training semantic segmentation models. In this work, we are the first to demonstrate that long-tailed distribution in training data can cause the CAM calculated through classifier weights over-activated for head classes and under-activated for tail classes due to the shared features among head- and tail- classes. This degrades pseudo-label quality and further influences final semantic segmentation performance.
To address this issue, we propose a \textbf{S}hared \textbf{F}eature \textbf{C}alibration (SFC) method for CAM generation. Specifically, we leverage the class prototypes that carry positive shared features and propose a \textbf{M}ulti-\textbf{S}caled \textbf{D}istribution-\textbf{W}eighted (MSDW) consistency loss for narrowing the gap between the CAMs generated through classifier weights and class prototypes during training. The MSDW loss counterbalances over-activation and under-activation by calibrating the shared features in head-/tail-class classifier weights.
Experimental results show that our SFC significantly improves CAM boundaries and achieves new state-of-the-art performances. The project is available at https://github.com/Barrett-python/SFC.
\end{abstract}
\section{Introduction}
Semantic segmentation \cite{minaee2021image} assigns semantic labels to image pixels and is crucial for applications like autonomous driving, robotics \cite{zhang2022mining}. Obtaining accurate pixel annotations for training deep learning models is laborious and time-consuming. 
One alternative approach is to adopt \textbf{W}eakly \textbf{S}upervised \textbf{S}emantic \textbf{S}egmentation (WSSS) with only image-level labels provided \cite{ahn2019weakly, wang2020self, zhang2020causal, lee2021anti, xu2022multi,zhang2023credible,zhang2021affinity}. Generally, these methods employ \textbf{C}lass \textbf{A}ctivation \textbf{M}apping (CAM) \cite{zhou2016learning} to generate discriminative semantic masks from a classification model. Then, a series of post-processing methods \cite{krahenbuhl2011efficient} are adopted to refine the masks to obtain pixel-level pseudo labels, which are then used to train a semantic segmentation model \cite{chen2017deeplab}. 

However, we find the training data of WSSS are naturally long-tailed distributed (Fig.~\ref{intro}(a)), which makes the shared feature components \cite{li2020group} tend to be positive in head-class classifier weight and negative in tail-class classifier weight because the head-class weight receives more positive gradients (denoted as $\oplus$) than the negative ones (denoted as $\ominus$) and the tail-class weight receives more negative gradients than the positive ones (Fig.~\ref{intro}(b)). This makes the pixels containing shared features activated by the head-class classifier weight (\emph{i.e.}, the dot product (denoted as $\cdot$) of feature and weight $>0$) while the pixels containing tail-class feature not activated by the tail-class weight (\emph{i.e.}, the dot product of feature and weight $<0$) as shown in Fig.~\ref{intro}(c). Thus, the CAM calculated through classifier weights inevitably becomes over-activated for head classes and under-activated for tail classes (Fig.~\ref{intro}(d)). 
This degrades the qualities of pseudo labels and further influences the final WSSS performances. On the other hand, as shown in Fig.~\ref{intro}(d), the CAM activated by the head-class prototype \cite{chen2022self} is less-activated compared to the CAM activated by the head-class classifier weight, and the CAM activated by the tail-class prototype is more-activated compared to the CAM activated by the tail-class classifier weight.

Inspired by the above findings (a detailed theoretical analysis is provided in \textbf{Analysis On SFC} section of main paper), we propose a \textbf{S}hared \textbf{F}eature \textbf{C}alibration (SFC) method to reduce shared feature proportions in head-class classifier weights and increase the ones in tail-class classifier weights, avoiding shared-feature-caused over-/under-activation issues. 
Particularly, a \textbf{M}ulti-\textbf{S}caled \textbf{D}istribution-\textbf{W}eighted (MSDW) consistency loss is calculated on the CAMs generated through class prototypes and classifier weights, where the consistency loss magnitude on one class is re-weighted by the total sample number gaps between this class and other classes. The theories behind this re-weighting strategy is also demonstrated, proving that pseudo-labels with better boundaries can be achieved through our SFC. 
The contributions of this work include:
\begin{itemize}
\item We first point out that the features shared by head and tail classes can enlarge the classifier-weight-generated CAM for the head class and shrink it for the tail class under a long-tailed scenario.
\item We propose a \textbf{S}hared \textbf{F}eature \textbf{C}alibration (SFC) CAM generation method aiming at balancing the shared feature proportions in different classifier weights, which can improve the CAM quality.
\item Our method achieves new state-of-the-art WSSS performances with only image-level labels on Pascal VOC 2012 and COCO 2014.

\end{itemize}
\section{Related Works}
\subsection{Weakly Supervised Semantic Segmentation} 
The generation of pseudo-labels in WSSS is based on attention mapping \cite{wang2020self,zhang2021affinity}. The key step is the produce of high-quality CAMs \cite{sun2020mining,yoon2022adversarial}. Several works have designed heuristic approaches, such as erasing and accumulation \cite{zhang2021complementary,yoon2022adversarial}, to force the network to mine novel regions rather solely focusing on discriminative regions. Moreover, other strategies include self-supervised learning \cite{wang2020self, chen2022self}, contrastive learning \cite{du2022weakly}, and cross-image information \cite{xu2023self} have been proposed to generate accurate and complete CAMs. Recently, vision-language pre-training has emerged as the prevalent approach for addressing downstream vision-language tasks \cite{zhu2023ctp}, including WSSS \cite{lin2023clip}. Due to the rough boundary of the initial map, refinement methods like CRF \cite{krahenbuhl2011efficient} and IRN \cite{ahn2019weakly} are employed for further enhancements. However, to the best of our knowledge, no previous work aims at solving the over-/under-activation issue caused by long-tailed distributed training data. This paper analyzes the reasons behind the over-/under-activation and tackles this issue through a \textbf{S}hared \textbf{F}eature \textbf{C}alibration (SFC) method.

\subsection{Shared Feature in Classification}
Classification is an upstream task for semantic segmentation \cite{zhang2023slca} and shared feature has been actively studied in this task \cite{li2020group}. Most existing methods \cite{zheng2017learning, yao2017autobd, peng2017object} tend to only extract discriminative partial features for classification and prevent the shared features from influencing the classification performance. Although both training under classification loss, unlike the classification task, WSSS cannot solely rely on discriminative features to construct intact CAM, and existing methods \cite{lee2022threshold,chen2022self} freeze several layers of pre-trained encoder for avoiding catastrophic forgetting of indiscriminative features \cite{vasconcelos2022proper}. In this work, we focus on balancing the shared feature proportions in classifier weights under a long-tailed scenario for a better WSSS performance.

\begin{figure*}[t!]
\centering
\includegraphics[height=4.6cm]{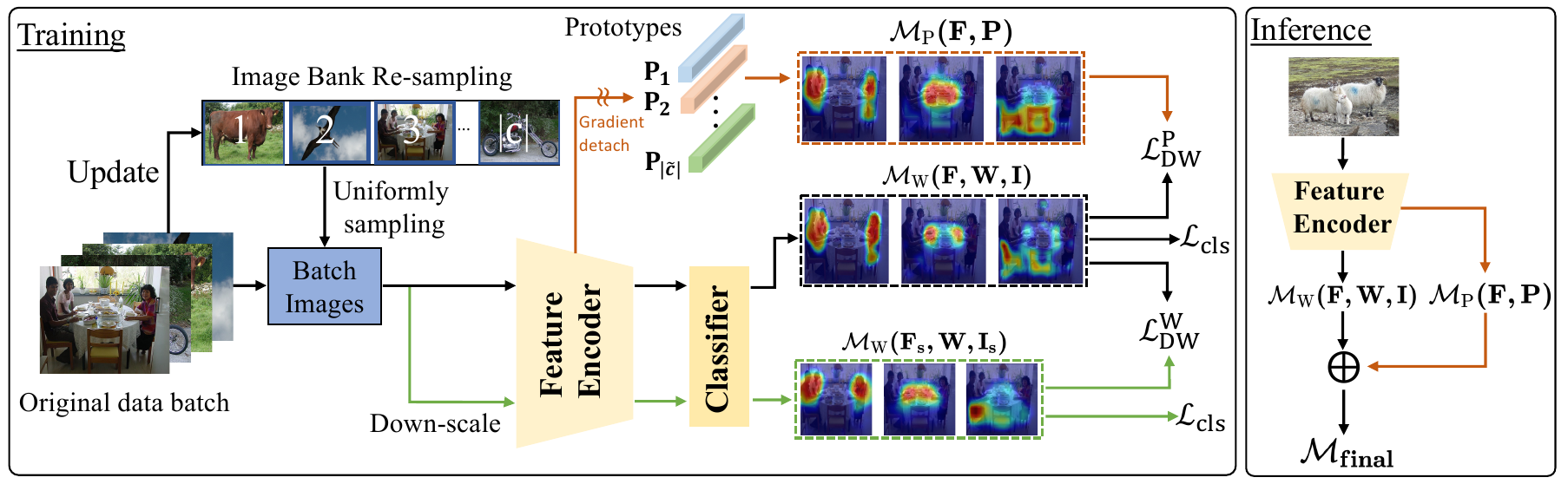}
\caption{The overall structure of our proposed SFC.
For each training image, two distribution-weighted consistency losses ($\mathcal{L}_{\text{DW}}^{\text{P}}$ and $\mathcal{L}_{\text{DW}}^{\text{W}}$) are calculated, where $\mathcal{L}_{\text{DW}}^{\text{P}}$ is calculated between the prototype CAM ($\mathcal{M}_{\text{P}}$) and classifier weight CAM ($\mathcal{M}_{\text{W}}$) of original image and $\mathcal{L}_{\text{DW}}^{\text{W}}$ is calculated between the classifier weight CAMs of down-scaled and original images. In addition, an image bank that stores the latest shown images for different classes is maintained, and images are uniformly sampled from it to complement the original training batch, increasing the consistency loss optimization frequency for tail classes.
Finally, the classifier weight CAM is complemented with prototype CAM in inference.}
\label{network}
\end{figure*}
\section{Methodology}
The pipeline of our SFC is illustrated in Fig.~\ref{network}. It involves an \textbf{I}mage \textbf{B}ank \textbf{R}e-sampling (IBR) and a \textbf{M}ulti-\textbf{S}caled \textbf{D}istribution-\textbf{W}eighted (MSDW) consistency loss. 
\subsection{Preliminary}
\subsubsection{Classifier Weight CAM}
Given an input image $\mathbf{I}$, the features extracted from $\mathbf{I}$ by the encoder is denoted as $\mathbf{F} \in \mathbb{R}^{H\times W \times D}$, and the classifier weight of class $c$ is denoted as $\mathbf{W}_c \in \mathbb{R}^{D \times 1}$, where $H\times W$ is the feature map size and $D$ is the feature dimension. The classification loss, which is a multi-label soft margin loss, is calculated as:
\begin{equation}
\begin{aligned}
\mathcal{L}_{\text{cls}}=&\frac{1}{\mathbf{\left| C \right|}} \sum_{c=1}^\mathbf{\left| C \right|} \bigg(\mathbf{y}_c \log \Big(sigmoid\big(GAP(\mathbf{F}\mathbf{W}_c)\big)\Big) \\
&+\left(1-\mathbf{y}_c\right) \log \Big(1-sigmoid\big(GAP(\mathbf{F}\mathbf{W}_c)\big)\Big)\bigg),
\end{aligned}
\end{equation}
where $\mathbf{C}$ is the foreground class set and $\mathbf{\left| C \right|}$ denotes its size; $\mathbf{y}_{c}$ denotes the binary label on class $c$; $GAP(\cdot)$ denotes the global average pooling. 
Then, the CAM generated through classifier weights $\mathbf{W}$ (\emph{i.e.}, classifier weight CAM) on the extracted feature $\mathbf{F}$ of input image $\mathbf{I}$ is calculated as follows: 

\begin{equation}
\begin{aligned}
    \mathbf{\mathcal{M}}_{\mathbf{W}}(\mathbf{F}, \mathbf{W},\mathbf{I})=&
    \underbrace{\left\{ {f}\left(\mathbf{F} \mathbf{W}_{1} ,\mathbf{I} ,\mathbf{F}\right),...,{f}\left(\mathbf{F} \mathbf{W}_{\left| \mathbf{C}\right|},\mathbf{I} ,\mathbf{F}\right) \right \}}_{\left| \mathbf{C}\right|\ foreground\ classes} \\
    &\cup \underbrace{\Big(\mathbf{1}- \max_{c \in \mathbf{C}} \big({f}(\mathbf{F} \mathbf{W}_{c},\mathbf{I} ,\mathbf{F})\big)\Big)}_{background \ class},
\end{aligned}
\label{pcmeq}
\end{equation}  
where $\mathbf{\mathcal{M}}_{\mathbf{W}}$ denotes the classifier weight CAM, $f(\cdot)$ denotes a function that feeds the normalized $ReLU$ activated $\mathbf{FW}$, $\mathbf{I}$, and $\mathbf{F}$ into a \textbf{P}ixel \textbf{C}orrelation \textbf{M}odule (PCM) \cite{wang2020self} to refine the CAM based on the relationships among the low-level features of different pixels.
\subsubsection{Prototype CAM}
Following \cite{chen2022self,chen2023saving}, class prototype is calculated through masked average pooling of extracted features. Specifically, the hierarchical features extracted from different layers of feature extractor are denoted as $\mathbf{F_1,F_2,F_3,F_4}$; $L(\cdot)$ denotes the linear projection and it stops gradients to the feature extractor. Then, the prototype $\mathbf{P}_{\tilde{c}}$ of class $\tilde{c}$ ($\tilde{c}$ can be either foreground or background class) is calculated as follows: 
\begin{equation}
\mathbf{P}_{\tilde{c}}= MAP\Big(\big(\widehat{\mathbf{\mathcal{M}}}_{\textbf{W}}(\mathbf{F},\mathbf{W},\mathbf{I})\big)_{\tilde{c}} \odot {L}\mathbf{(F_1,F_2,F_3,F_4)}\Big),
\end{equation}  
where $\big(\widehat{\mathbf{\mathcal{M}}}_{\textbf{W}}(\mathbf{F}, \mathbf{W},\mathbf{I})\big)_{\tilde{c}}$ is a binary mask for class $\tilde{c}$, highlighting the pixels whose activation values are higher than the set threshold with $1$; $MAP(\cdot)$ denotes masked average pooling. Finally, the CAM calculated through the prototype of class $\tilde{c}$ (\emph{i.e.}, prototype CAM) is calculated as follows:
\begin{equation}
\big(\mathbf{\mathcal{M}}_{\textbf{P}}\mathbf{(F,P)}\big)_{\tilde{c}}= ReLU\big(\cos \left \langle   \mathbf{P}_{\tilde{c}}, L \mathbf{(F_1,F_2,F_3,F_4)} \right \rangle \big),
\end{equation} 
where $\cos \left \langle  \cdot,\cdot \right \rangle $ denotes the cosine similarity between the two terms within it.
\subsection{Shared Feature Calibration}
\subsubsection{Image Bank Re-sampling (IBR)}\label{31}
We maintain an image bank $\mathcal{B} = (\mathbf{b}_{1},...,\mathbf{b}_{c})$ that stores $\mathbf{\left| C \right|}$ images for $\mathbf{\left| C \right|}$ foreground classes. For each image $\mathbf{I}$ in the current training batch, we update $\mathbf{b}_{c}$ with $\mathbf{I}$ when the $c$-th class appears in $\mathbf{I}$. Otherwise, we keep $\mathbf{b}_c$ as it was. After the bank updating, we uniformly sample $N_{\text{IBR}}$ images from the current bank and concatenate them with the original training batch as final training inputs. The uniform sampling does not bring further shared feature issues caused by long-tailed distribution, as the sample numbers of different classes are nearly balanced. 

Our proposed IBR is used for increasing the tail-class sampling frequency, thus the MSDW loss will be enforced on the tail classes more frequently, effectively calibrating the shared features in the tail-class classifier weights.

\subsubsection{Multi-Scaled Distribution-Weighted Consistency Loss}
\label{MSDW_method}
To address the over-activation issues on head classes and under-activation issues on tail classes, we propose two \textbf{D}istribution-\textbf{W}eighted (DW) consistency losses $\mathcal{L}_{\text{DW}}^{\text{P}}$ and $\mathcal{L}_{\text{DW}}^{\text{W}}$.
$\mathcal{L}_{\text{DW}}^{\text{P}}$ is calculated between the prototype CAM and classifier weight CAM as: 
\begin{equation}
\begin{aligned}
\mathcal{L}_{\text{DW}}^{\text{P}} =& \textstyle{\sum_{c=1}^{|\mathbf{C}|}}\Big({DC}_c \cdot \underbrace{\big\|\big(\mathbf{\mathcal{M}}_{\mathbf{W}}\big)_{c}-\big(\mathbf{\mathcal{M}}_{\mathbf{P}}\big)_{c}\big\|_1}_{\ell1 \ loss \ of \ foreground \ class }\Big)\\&+\underbrace{\big\|\big(\mathbf{\mathcal{M}}_{\mathbf{W}}\big)_{|\mathbf{C}|+1}-\big(\mathbf{\mathcal{M}}_{\mathbf{P}}\big)_{|\mathbf{C}|+1}\big\|_1}_{\ell1 \ loss \ of \ background \ class }\ 
,
\end{aligned}
\label{eqsdb}
\end{equation}
where $DC_c$ denotes the scaled \textbf{D}istribution \textbf{C}oefficient, calculated for each foreground class $c$ as:
\begin{equation}
    \begin{aligned}
    {DC}_{c} = &\underbrace{\frac{|\mathbf{C}|} {\textstyle{\sum_{j=1}^{|\mathbf{C}|}} \big( \frac{ \textstyle{\sum_{i=1}^{|\mathbf{C}|}} |n_j - n_i|}{(n_j+ \mathcal{N})}  \big) }}_{scaling \ factor} \cdot\underbrace{\frac{\textstyle{\sum_{i=1}^{|\mathbf{C}|}} |n_c - n_i|}{n_c+\mathcal{N}}}_{total \ demand} , 
    \end{aligned} 
\label{ROCeq}
\end{equation}
where $n_c$ denotes the sample number of class $c$; $\mathcal{N}$ denotes the estimated increased sample number of our IBR on each class and $\mathcal{N} =\frac{N_{\text{IBR}}\cdot C_{\text{IBR}}\cdot N_{\text{iter}}}{|\mathbf{C}|}$. Here, $N_{\text{iter}}$ denotes the number of iterations in one training epoch; $N_{\text{IBR}}$ is the sampling number from the image bank; $C_{\text{IBR}}$ is the average number of classes covered in one image.

We calculate the sum of sample number gaps between class $c$ and other classes and regard this sum as the \emph{total demand} on the consistency loss for class $c$. Next, this total reward is averaged by $n_c+\mathcal{N}$ (\emph{i.e.}, the estimated total sample number of class $c$) and scaled with the \emph{scaling factor}, obtaining the scaled distribution coefficient (\emph{i.e.}, ${DC}_{c}$).
The \emph{scaling factor} is to scale the $\ell$1 loss magnitude of foreground class to the same level as the background class.

${DC}_{c}$ is finally used to re-weight the consistency loss on class $c$, assigning higher consistency loss to the class with higher total demand, as the severity of over-/under-activation issue is positively related to the \emph{total demand}.

Meanwhile, all images in the current training batch are further down-scaled with 0.5 through bilinear interpolation algorithm and are used to calculate the loss $\mathcal{L}_{\text{DW}}^{\text{W}}$: 
\begin{equation}
\begin{aligned}
\mathcal{L}_{\text{DW}}^{\text{W}} = & \textstyle{\sum_{c=1}^{|\mathbf{C}|}}\bigg({DC}_c \cdot \Big\|s\Big(\big(\mathbf{\mathcal{M}}_{\mathbf{W}}(\mathbf{F},\mathbf{W},\mathbf{I})\big)_{c}\Big) \\
&-\big(\mathbf{\mathcal{M}}_{\mathbf{W}}(\mathbf{F}_{s},\mathbf{W},\mathbf{I}_{s})\big)_{c}\Big\|_1\bigg),
\end{aligned}
\label{eqsdb2}
\end{equation}
where $s(\cdot)$ denotes the bilinear down-sampling operation; $\mathbf{I}_{s}$ denotes the down-scaled image and $\mathbf{F}_{s}$ denotes its extracted feature. Similar to Eq.~\eqref{eqsdb}, we re-weight the consistency loss by $DC$ coefficient. Considering the prototype CAM on the down-scaled image is less accurate than the down-scaled classifier weight CAM calculated on the original image (see \textbf{$\mathcal{L}^{\text{W}}_{\text{DW}}$ with Multi-Scaled Scheme} in appendix), we calculate the consistency loss between the down-scaled classifier weight CAM on the original image and the classifier weight CAM on the down-scaled image. $\mathcal{L}_{\text{DW}}^{\text{W}}$ further boosts the performance improvement. Our multi-scaled distribution-weighted consistency loss $\mathcal{L}_{\text {MSDW}}$ is formulated  as follows:
\begin{equation}
\mathcal{L}_{\text {MSDW}}=\mathcal{L}_{\text{cls}}+\mathcal{L}_{\text{DW}}^{\text{P}}+\mathcal{L}_{\text{DW}}^{\text{W}}.
\end{equation}

\subsubsection{Inference}\label{33}
The final CAM for inference is calculated as:
\begin{equation}
(\mathbf{\mathcal{M}}_{\text{final}})_{\tilde{c}} = \big(\mathbf{\mathcal{M}}_{\mathbf{W}} (\mathbf{F},\mathbf{W},\mathbf{I})\big)_{\tilde{c}}+ \\ \big(\mathbf{\mathcal{M}}_{\mathbf{P}}(\mathbf{F},\mathbf{P})\big)_{\tilde{c}},
\label{eqinfer}
\end{equation}
where $(\mathbf{\mathcal{M}}_{\text{final}})_{\tilde{c}}$ denotes the final CAM of class $\tilde{c}$; $\tilde{c}$ can be foreground or background class. In this way, the classifier weight CAM is complemented by the prototype CAM, jointly solving the over-/under-activation issue. 

\section{Analysis on SFC}
This section demonstrates how shared features in classifier weights cause over-/under-activation issues under a long-tailed scenario and the working mechanism behind SFC.

\subsection{Shared Feature Distribution in Classifier Weights}
\label{sfinclassifier}
In image-level WSSS, a multi-label soft margin loss (denoted as $\mathcal{L}$) is commonly used for the classification model training \cite{chen2022self}. Following the definitions in \cite{tan2021equalization}, the positive and negative gradients caused by $\mathcal{L}$ are formulated as follows:
\begin{equation}
    \label{first}
    \begin{split}
        \begin{aligned}
            (\mathcal{L}'^{pos}_{c})_i =&-\mathbf{y}_{i,c}\big(sigmoid(\mathbf{z}_{i,c})-1\big)>0 ,  \quad  \mathbf{y}_{i,c}=1
             \end{aligned}\\
             \begin{aligned}
            (\mathcal{L}'^{neg}_{c})_i  =&-(1-\mathbf{y}_{i,c})sigmoid(\mathbf{z}_{i,c})<0, \quad \mathbf{y}_{i,c}=0,
       \end{aligned}
    \end{split}
\end{equation}
where $\mathbf{z}_{i,c}$ denotes the model predicted logit for the $i$-th sample on class $c$, and $sigmoid(\mathbf{z}_{i,c})$ indicates its sigmoid activated value; $\mathbf{y}_{i,c}$ denotes the label of the $i$-th sample on class $c$ (either 0 or 1); 
$(\mathcal{L}'^{pos}_{c})_i$ and $(\mathcal{L}'^{neg}_{c})_i$ denote the positive and negative gradients on $\mathbf{z}_{i,c}$.

Considering a simplified case with one head class $H$ and one tail class $T$, based on the conclusion that different classes have shared features \cite{hou2022batchformer,liu2021infrared,li2020group}, the head-class image feature can be decomposed as $\alpha_{H}  \mathbf{f}_{H}+\eta_{H} \mathbf{f}_{0}$, where ${\alpha_{H}}  \mathbf{f}_{H}$ and ${\eta_{H}} \mathbf{f}_{0}$ indicate the discriminative and shared feature parts respectively, with ${\alpha_{H}}$ and ${\eta_{H}}$ indicating their proportions. Similarly, the tail-class image feature can be decomposed as $\alpha_{T} \mathbf{f}_{T}+\eta_{T}  \mathbf{f}_{0}$.
Then, the head-class classifier weight $\mathbf{W}_{H}$ can be represented as:
\begin{equation}
    \label{eq11}
    \begin{aligned}
        \mathbf{W}_{H}
        =& n_{H}  E(\alpha_{H})E(\mathcal{L}'^{pos}_{H}) \mathbf{f}_{H}+ n_{T}E(\alpha_{T})  E(\mathcal{L}'^{neg}_{H}) \mathbf{f}_{T}\\&+\underbrace{\big(n_{H} E(\eta_{H}) E(\mathcal{L}'^{pos}_{H}) +n_{T} E(\eta_{T}) E(\mathcal{L}'^{neg}_{H})\big) \mathbf{f}_{0}}_{\mathbf{f}_{0}^{H}: \ shared \ feature \ component \ in \ \mathbf{W}_{H}},       
    \end{aligned}
\end{equation}
where $n_H$ and $n_T$ indicate the sample numbers of head and tail classes, with $n_{H} \gg n_{T}$ under a long-tailed scenario; $E(\cdot)$ denotes the expectation operation. Similarly, we have the tail-class classifier weight $\mathbf{W}_{T}$ as:
\begin{equation}
    \label{eq2}
    \begin{aligned}
    \mathbf{W}_{T}=&n_{T} E(\alpha_{T})E(\mathcal{L}'^{pos}_{T}) \mathbf{f}_{T} + n_{H} E(\alpha_{H})E(\mathcal{L}'^{neg}_{T}) \mathbf{f}_{H}\\&+\underbrace{\big(n_{T} E(\eta_{T}) E(\mathcal{L}'^{pos}_{T})+n_{H} E(\eta_{H}) E(\mathcal{L}'^{neg}_{T})\big) \mathbf{f}_{0}}_{\mathbf{f}_{0}^{T}: \ shared \ feature \ component \ in \ \mathbf{W}_{T}}.
    \end{aligned}
\end{equation}
The proofs of Eq.~\eqref{eq11} and Eq.~\eqref{eq2} are provided in \textbf{Proof 1} of appendix.

Then, as demonstrated in \textbf{Gradient Magnitude Analysis} of appendix, the magnitude of $E(\mathcal{L}'^{pos})$ is larger than that of $E(\mathcal{L}'^{neg})$ and the gap is not significant. Combining with the precondition that $n_{H} \gg n_{T}$,
it can be concluded that in Eq.~\eqref{eq11} we have $n_{H} E(\mathcal{L}'^{pos}_{H})+n_{T} E(\mathcal{L}'^{neg}_{H})>0$. Similarly, in Eq.~\eqref{eq2} we have   $n_{T}E(\mathcal{L}'^{pos}_{T})+n_{H}E(\mathcal{L}'^{neg}_{T}) <0$.
Then, based on Eq.~\eqref{eq11} and Eq.~\eqref{eq2}, we can have:
\\\\
\noindent\textbf{Conclusion 1.} \textit{When $n_{H} \gg n_{T}$ and the difference between $E(\eta_{H})$ and $E(\eta_{T})$ is not as significant as that between $n_{H}$ and $n_{T}$, the shared feature component in $\mathbf{W}_{H}$ (\emph{i.e.}, $\mathbf{f}_{0}^{H}$) tends to be positive with a large magnitude, and the one in $\mathbf{W}_{T}$ (\emph{i.e.}, $\mathbf{f}_{0}^{T}$) tends to be negative with a large magnitude. However, when $n_{H} \approx n_{T}$, the class with a higher $E(\eta)$ will have a larger shared feature in its classifier weight, and the shared feature magnitude will be much lower than that when $n_{H} \gg n_{T}$.}

\subsection{Over-activation and Under-activation}
\label{overunderact}
One extracted feature of tail-class image area can be decomposed as $\mathbf{A}_{T} = \alpha_{T}^{A}\mathbf{f}_{T}+\eta_{T}^{A}\mathbf{f}_{0}$ ($\alpha_{T}^{A}$ and $\eta_{T}^{A}$ are the proportions of $\mathbf{f}_{T}$ and $\mathbf{f}_{0}$). Similarly, one extracted feature of head-class image area can be decomposed as $\mathbf{A}_{H} = \alpha_{H}^{A}\mathbf{f}_{H}+\eta_{H}^{A}\mathbf{f}_{0}$. Under long-tailed scenario (\emph{i.e.}, $n_{H} \gg n_{T}$), the head-/tail-class classifier weight activations on $\mathbf{A}_{T}$ and $\mathbf{A}_{H}$ can be formulated as follows:
\begin{equation}
    \label{eq12}
    \begin{aligned}
&\mathbf{A}_{T}
\mathbf{W}_{H}\\= &{\big(n_{H} E(\eta_{H}) E(\mathcal{L}'^{pos}_{H})  
+n_{T} E(\eta_{T}) E(\mathcal{L}'^{neg}_{H})\big) \eta_{T}^{A}\left \| \mathbf{f}_{0} \right \|^{2} _{2}}\\ 
&+n_{T}\alpha_{T}^{A}E(\alpha_{T})E(\mathcal{L}'^{neg}_{H})\left \| \mathbf{f}_{T} \right \|^{2} _{2},
    \end{aligned}
\end{equation}
\begin{equation}
    \label{eq12_ahwh}
    \begin{aligned}
    &\mathbf{A}_{H} \mathbf{W}_{H}\\=& \big(n_{H} E(\eta_{H}) E(\mathcal{L}'^{pos}_{H})+n_{T} E(\eta_{T}) E(\mathcal{L}'^{neg}_{H})\big) \eta_{H}^{A}\left \| \mathbf{f}_{0} \right \|^{2} _{2}\\ &+n_{H}\alpha_{H}^{A}E(\alpha_{H})E(\mathcal{L}'^{pos}_{H})\left \| \mathbf{f}_{H} \right \|^{2} _{2},
    \end{aligned}   
\end{equation}
\begin{equation}
    \label{eq12_2}
    \begin{aligned}
    &\mathbf{A}_{T} \mathbf{W}_{T}\\=& \big(n_{T} E(\eta_{T}) E(\mathcal{L}'^{pos}_{T}) +n_{H} E(\eta_{H}) E(\mathcal{L}'^{neg}_{T})\big)\eta_{T}^{A} \left \| \mathbf{f}_{0} \right \|^{2} _{2}\\ &+n_{T}\alpha_{T}^{A}E(\alpha_{T})E(\mathcal{L}'^{pos}_{T})\left \| \mathbf{f}_{T} \right \|^{2} _{2},
    \end{aligned}   
\end{equation}
\begin{equation}
    \label{eq12_ahwt}
    \begin{aligned}
    &\mathbf{A}_{H} \mathbf{W}_{T}\\=& \big(n_{T} E(\eta_{T}) E(\mathcal{L}'^{pos}_{T}) +n_{H} E(\eta_{H})E(\mathcal{L}'^{neg}_{T})\big)\eta_{H}^{A} \left \| \mathbf{f}_{0} \right \|^{2} _{2}\\
&+n_{H}\alpha_{H}^{A}E(\alpha_{H})E(\mathcal{L}'^{neg}_{T})\left \| \mathbf{f}_{H} \right \|^{2} _{2}.
    \end{aligned}   
\end{equation}
Based on \textbf{Conclusion 1}, proved through \textbf{Proof 2} in appendix, we have \textbf{Conclusion 2}:\\\\
\noindent\textbf{Conclusion 2.} \textit{When $n_{H} \gg n_{T}$, $\mathbf{A}_{H}\mathbf{W}_{T}$ and $\mathbf{A}_{T}\mathbf{W}_{T}$ tend to be unactivated, and $\mathbf{W}_{T}$ has \textbf{under-activated} tail-class image area compared with the ground truth (as shown in the tail classes of Fig.~\ref{fig_analysis}(b)). On the contrary, $\mathbf{A}_{H}\mathbf{W}_{H}$ and $\mathbf{A}_{T}\mathbf{W}_{H}$ tend to be activated, and $\mathbf{W}_{H}$ has \textbf{over-activated} head-class image area compared with the ground truth (as shown in the head classes of Fig.~\ref{fig_analysis}(b)).}
\\\\
On the other hand, for the class prototype extracted through averaging its classifier weight activated features, it only has positive shared features. Thus, based on \textbf{Conclusion 2}, proved through \textbf{Proof 3} in appendix, we have \textbf{Conclusion 3}: 
\\\\
\noindent\textbf{Conclusion 3.} \textit{Let $\mathbf{P}_{H}$ and $\mathbf{P}_{T}$ denote the prototypes of head class $H$ and tail class $T$, $\mathbf{A}$ denotes the image area including $\mathbf{A}_{H}$ and $\mathbf{A}_{T}$. When $n_{H} \gg n_{T}$,
$\mathbf{A}\mathbf{P}_{H}$ is \textbf{less-activated} compared with $\mathbf{A}\mathbf{W}_{H}$ (as shown in the head classes of Fig.~\ref{fig_analysis}(b) and Fig.~\ref{fig_analysis}(c)). On the contrary, $\mathbf{A}\mathbf{P}_{T}$ is \textbf{more-activated} compared with $\mathbf{A}\mathbf{W}_{T}$ (as shown in the tail classes of Fig.~\ref{fig_analysis}(b) and Fig.~\ref{fig_analysis}(c)).}

\subsection{How SFC Works}
\label{howsdbworks}
As described in Eq.~\eqref{eqsdb}, our DW consistency loss pulls closer prototype CAM and classifier weight CAM for pairs: $\left \{\mathbf{A}_{T}\mathbf{W}_H,\mathbf{A}_{T}\mathbf{P}_H\right \}$, $\left \{\mathbf{A}_{T}\mathbf{W}_T,\mathbf{A}_{T}\mathbf{P}_T\right \}$, $\left \{\mathbf{A}_{H}\mathbf{W}_H ,\mathbf{A}_{H}\mathbf{P}_H \right \}$, $\left \{\mathbf{A}_{H}\mathbf{W}_T,\mathbf{A}_{H}\mathbf{P}_T\right \}$. Thereby, $\mathbf{W}_H$ or $\mathbf{W}_T$ are enforced to learn towards features activated by $\mathbf{P}_H$ or $\mathbf{P}_T$.
When $n_{H} \gg n_{T}$, based on \textbf{Conclusion 3}, we have:

\textbf{CASE 1}: For $\left \{ \mathbf{A}_{T}\mathbf{W}_H,\mathbf{A}_{T}\mathbf{P}_H \right \}$, $\mathbf{A}_{T}\mathbf{P}_H$ is \textbf{less-activated} compared with $\mathbf{A}_{T}\mathbf{W}_H$. Since $\mathbf{A}_{T}$ contains $\mathbf{f}_{0}$ and $\mathbf{f}_{T}$,  $\mathbf{W}_{H}$ is optimized towards $-\mathbf{f}_{0}$ and $-\mathbf{f}_{T}$, bringing \emph{positive} effects for $\mathbf{W}_H$ to shrink its CAM on tail-class areas. 

\textbf{CASE 2}: For $\left \{\mathbf{A}_{T}\mathbf{W}_T,\mathbf{A}_{T}\mathbf{P}_T\right \}$, $\mathbf{A}_{T}\mathbf{P}_T$ is \textbf{more-activated} compared with $\mathbf{A}_{T}\mathbf{W}_T$. Since  
$\mathbf{A}_{T}$ contains $\mathbf{f}_{0}$ and $\mathbf{f}_{T}$, $\mathbf{W}_{T}$ is optimized towards $\mathbf{f}_{0}$ and $\mathbf{f}_{T}$, bringing \emph{positive} effects for $\mathbf{W}_T$ to expand its CAM on tail-class areas. 

\textbf{CASE 3}: For $\left \{\mathbf{A}_{H}\mathbf{W}_T, \mathbf{A}_{H}\mathbf{P}_T\right \}$, $\mathbf{A}_{H}\mathbf{P}_T$ is \textbf{more-activated} compared with $\mathbf{A}_{H}\mathbf{W}_T$. Since $\mathbf{A}_{H}$ contains $\mathbf{f}_{0}$ and $\mathbf{f}_{H}$, $\mathbf{W}_{T}$ is optimized towards $\mathbf{f}_{0}$ and $\mathbf{f}_{H}$. As $\mathbf{W}_{T}$ has $-\mathbf{f}_{0}$ and $-\mathbf{f}_{H}$ with large magnitudes (\textbf{Conclusion 1}), optimizing $\mathbf{W}_{T}$ towards positive $\mathbf{f}_{0}$ and $\mathbf{f}_{H}$ hardly brings negative effects. 

\textbf{CASE 4}: For $\left \{\mathbf{A}_{H}\mathbf{W}_H,\mathbf{A}_{H}\mathbf{P}_H\right \}$, $ \mathbf{A}_{H}\mathbf{P}_H$ is \textbf{less-activated} compared with $\mathbf{A}_{H}\mathbf{W}_H$. Since $\mathbf{A}_{H}$ contains $\mathbf{f}_{0}$ and $\mathbf{f}_{H}$, $\mathbf{W}_{H}$ is optimized towards $-\mathbf{f}_{0}$ and $-\mathbf{f}_{H}$. As $\mathbf{W}_{H}$ has $\mathbf{f}_{0}$ and $\mathbf{f}_{H}$ with large magnitudes (\textbf{Conclusion 1}), optimizing $\mathbf{W}_{H}$ towards $-\mathbf{f}_{0}$ and $-\mathbf{f}_{H}$ hardly brings negative effects. 

\begin{figure*}[t!]
\centering
    \includegraphics[height=5cm]{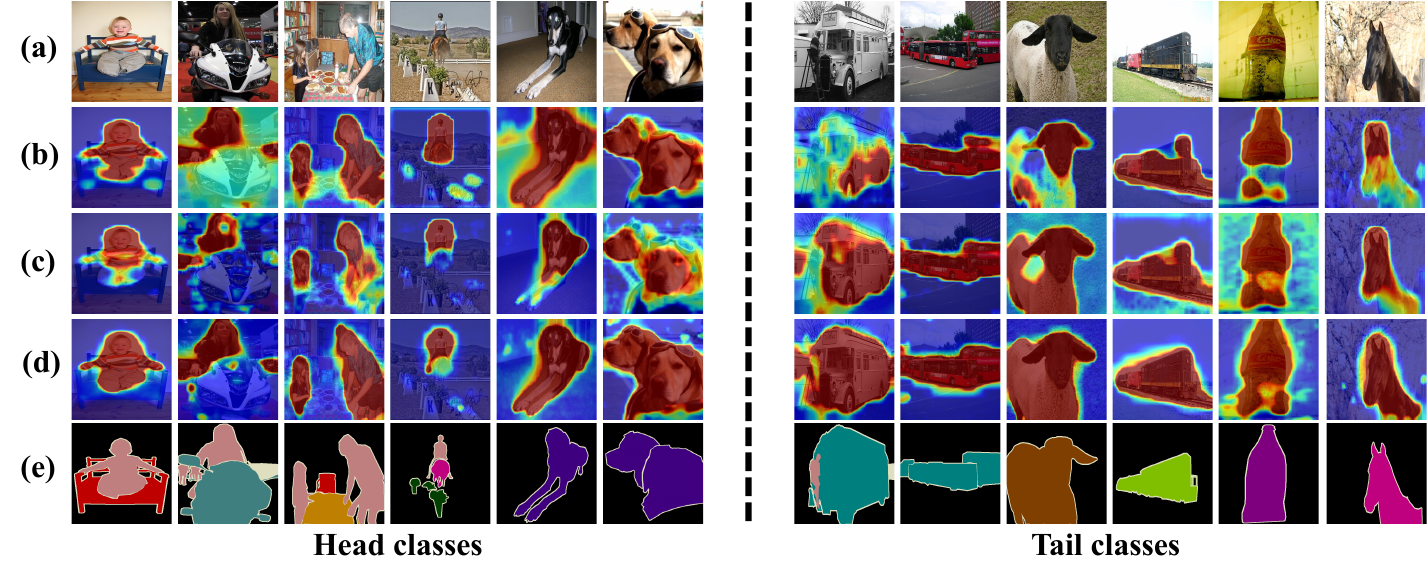}
    \caption{CAM visualization results on PASCAL VOC 2012, demonstrating \textbf{Conclusion 2} and \textbf{Conclusion 3}. (a) input images; (b) classifier weight CAMs; (c) prototype CAMs; (d) final CAMs generated through our SFC; (e) ground truth.}
    \label{fig_analysis}
\end{figure*}

In summary, classifier weights with severe over-/under-activation issues can benefit from \textbf{CASE 1} and \textbf{CASE 2}, while they are not negatively affected for \textbf{CASE 3} and \textbf{CASE 4}, improving
the overall CAMs as shown in Fig.~\ref{fig_analysis}(d).

However, when $n_{H} \approx n_{T}$, the consistency \emph{negatively} affects the CAM generation. For example, by pulling closer the pair of $\left \{\mathbf{A}_{H}\mathbf{W}_T, \mathbf{A}_{H}\mathbf{P}_T\right \}$, as $\mathbf{P}_T$ contains $\mathbf{f}_{0}$ and it activates $\mathbf{A}_{H}$ which contains $\mathbf{f}_{0}$ and $\mathbf{f}_{H}$, $\mathbf{W}_{T}$ will be optimized towards $\mathbf{f}_{0}$ and $\mathbf{f}_{H}$. However, $\mathbf{W}_{T}$ does not have $-\mathbf{f}_{0}$ or $-\mathbf{f}_{H}$ with large magnitude when $n_{H} \approx n_{T}$ (\textbf{Conclusion 1}), increasing $\mathbf{f}_{H}$ and $\mathbf{f}_{0}$ in $\mathbf{W}_{T}$ brings a \emph{negative} effect. 

Considering the consistency loss brings \emph{positive} effects when $n_{H} \gg n_{T}$ and brings \emph{negative} effects when $n_{H} \approx n_{T}$, 
we define the total demand on the consistency loss for each class by adding up the sample number gaps between this class and all other classes, and then regard this total demand as the weight of consistency loss on this class (\emph{i.e.}, $DC$ coefficient in Eq.~\eqref{eqsdb}), maximizing the consistency loss effect.
\section{Experiments}
\noindent\textbf{Dataset and Evaluation Metric.} Experiments are conducted on two benchmarks: PASCAL VOC 2012 \cite{everingham2010pascal} with 21 classes and MS COCO 2014 \cite{lin2014microsoft} with 81 classes. For PASCAL VOC 2012, following \cite{wang2020self,lee2021anti,chen2022self,li2022expansion}, we use the augmented SBD set \cite{hariharan2011semantic} with 10,582 annotated images. \textbf{M}ean \textbf{I}ntersection \textbf{o}ver \textbf{U}nion (mIoU) \cite{long2015fully} is used to evaluate segmentation results.

\noindent\textbf{Implementation Details.} For pseudo label generation, we adopt the ImageNet \cite{deng2009imagenet} pretrained ResNet50 \cite{he2016deep}. Random cropping size 512$\times$512 is adopted for training data augmentation. The $N_{\text{IBR}}$ is set to 4. $\mathcal{M}_{final}$ from our method is further post-processed by DenceCRF \cite{krahenbuhl2011efficient} and IRN \cite{ahn2019weakly} to generate the final pseudo labels, which are used to train the segmentation model: ResNet101-based DeepLabV2 \cite{ahn2019weakly,chen2022self}. More details can be found in appendix.

\subsection{Comparison of Pseudo-label Quality} 
\begin{table}[t!]
\centering
\small
\begin{tabular}{lccccc}
\toprule[1pt]
\multirow{3}*{Method} & \multicolumn{3}{c}{PASCAL VOC} \\ 
\cmidrule(r){2-4}
~ & CAM  & CRF & Mask  \\ 
\midrule 
IRN \cite{ahn2019weakly} &  48.8 & 54.3 & 66.3  \\
SEAM \cite{wang2020self} & 55.4 & 56.8 & 63.6 \\
CONTA \cite{zhang2020causal} & 48.8 & - & 67.9 \\
AdvCAM \cite{lee2021anti} & 55.6  & 62.1 &  68.0  \\
RIB \cite{lee2021reducing}& 56.5 & 62.9 &70.6\\
CLIMS \cite{xie2022clims}&56.6  &  62.4 & 70.5 \\
ESOL \cite{li2022expansion} & 53.6 & 61.4 & 68.7 \\ 
SIPE \cite{chen2022self} & 58.6 &64.7 & 68.0\\ 
AMN \cite{lee2022threshold} & 62.1  & 65.3 & 72.2\\ 
\midrule
 \textbf{SFC (Ours)} &\textbf{64.7} & \textbf{69.4} & \textbf{73.7}\\
\bottomrule
\end{tabular}
\caption{Evaluation (mIoU (\%)) of different pseudo labels on PASCAL VOC 2012 training set.}
\label{labelVOC}
\end{table}

To validate the effectiveness of our SFC, we evaluate the quality of intermediate and final results in the pseudo-label generation process in Table \ref{labelVOC}. Specifically, we first compare the initial CAM generated by the classification model (denoted as CAM). Then, we compare various post-processed CAMs to show the consistent improvements by our SFC. Particularly, the original CAM is first refined by CRF \cite{krahenbuhl2011efficient}(denoted as CRF) and further processed by IRN \cite{ahn2019weakly} to generate the final pseudo masks (denoted as Mask). Experimental results in Table \ref{labelVOC} show that the CAM from SFC is significantly better than the previous works on datasets with different class numbers and long-tailed degrees, and our method outperforms state-of-the-art methods by 2.6\% on PASCAL VOC. Regarding the CRF-post-processed CAM, we achieve 69.4\% mIoU on the PASCAL VOC, and further with IRN, our SFC improves the mIoU to 73.7\%, achieving 1.5\% gain compared to AMN \cite{lee2022threshold}.

\subsection{Comparison of WSSS Performance}
In WSSS, the CRF and IRN post-processed pseudo masks obtained from the initial CAM are treated as ground truth to train semantic segmentation model in a fully supervised manner. Table \ref{labelVOCfinal} reports the mIoU scores of our method and recent WSSS methods on the validation and test sets of PASCAL VOC 2012. On this dataset, we achieve 71.2\% and 72.5\% mIoU using an ImageNet pre-trained backbone, outperforming all other WSSS methods that use only image-level labels or both image-level labels and saliency maps \cite{xu2021leveraging, yao2021non,lee2021railroad}. Table \ref{labelcoco} reports the performance comparison on MS COCO 2014. Using the same training scheme as the PASCAL VOC 2012 experiment, our method achieves 46.8\% mIoU on the validation set with a ResNet101 backbone, outperforming AMN \cite{lee2022threshold} by 2.1\%.

\begin{table*}[t!]
\begin{minipage}[t]{0.5\textwidth}
\makeatletter\def\@captype{table}
\centering
\setlength{\tabcolsep}{0.5mm}
\begin{tabular}{lccc}
\toprule[1pt]
Method & Backbone & Val & Test \\ 
\midrule 
\multicolumn{3}{l}{\textbf{Image-level supervision + Saliency maps.}} \\
AuxSegNet \cite{xu2021leveraging} &  ResNet38 & 69.0 &68.6 \\ 
NSROM \cite{yao2021non} &  ResNet101 & 70.4 &70.2 \\
EPS \cite{lee2021railroad}  & ResNet101  & 71.0 &  71.8  \\
\midrule  
\multicolumn{3}{l}{\textbf{Image-level supervision only.}} \\
SEAM \cite{wang2020self}  &ResNet38  &  64.5 & 65.7 \\
PPC+SEAM \cite{du2022weakly}   &ResNet38 & 67.7 & 67.4 \\ 
ReCAM \cite{chen2022class}  &ResNet38 & 68.5 & 68.4 \\
SIPE \cite{chen2022self} &ResNet38 &68.2 &69.5 \\ 
SIPE \cite{chen2022self}  & ResNet101 & 68.8 &69.7\\ 
ESOL \cite{li2022expansion}  &ResNet101 & 69.9 & 69.3 \\ 
AMN \cite{lee2022threshold} &ResNet101 & 70.7  & 70.6  \\ \midrule 
\textbf{SFC (Ours)} & ResNet38 &\textbf{70.2} & \textbf{${\textbf{71.4}}$}  \\ \vspace{-0.05cm}
\textbf{SFC (Ours)} & ResNet101 &\textbf{71.2}  & \textbf{$\textbf{72.5}$} \\
\bottomrule
\end{tabular}
\caption{Comparison of semantic segmentation performance on PASCAL VOC 2012 validation and test sets.}
\label{labelVOCfinal}
\end{minipage}
~~~~
\begin{minipage}[t]{0.45\textwidth}
\makeatletter\def\@captype{table}
\centering
\setlength{\tabcolsep}{0.5mm}
\begin{tabular}{lcc}
\toprule[1pt]
Method & Backbone  & Val \\ 
\midrule 
\multicolumn{3}{l}{\textbf{Image-level supervision + Saliency maps.}} \\
G-WSSS \cite{li2021group}  & ResNet38  & 28.4 \\ 
AuxSegNet \cite{xu2021leveraging} & ResNet38 & 33.9 \\
EPS \cite{lee2021railroad}  & ResNet101 & 35.7 \\
\midrule
\multicolumn{3}{l}{\textbf{Image-level supervision only.}} \\
IRN \cite{ahn2019weakly} & ResNet50 & $32.6$ \\
SEAM \cite{wang2020self}  & ResNet38 & $31.9$ \\
RIB \cite{lee2021reducing} & ResNet38 & $43.8$ \\
SIPE \cite{chen2022self}  & ResNet38 & $43.6$ \\
CONTA \cite{zhang2020causal} & ResNet101 & $33.4$ \\
SIPE \cite{chen2022self}  & ResNet101 & $40.6$ \\
ESOL \cite{li2022expansion}  & ResNet101 & $42.6$ \\
AMN \cite{lee2022threshold}  & ResNet101 & $44.7$ \\
 \midrule 
\textbf{SFC (Ours)} & ResNet101 &\textbf{46.8} \\
\bottomrule
\end{tabular}
\caption{Comparison of semantic segmentation performance on MS COCO 2014 validation set.}
\label{labelcoco}
\end{minipage}
\end{table*}

\begin{table*}[t!]
\begin{minipage}[t]{0.32\textwidth}
\makeatletter\def\@captype{table}
\centering
\setlength{\tabcolsep}{0.4mm}
\begin{tabular}{ccccc}
\toprule  
 &IBR \ & $\mathcal{L}_{\text{DW}}^{\text{P}}$ \ & $\mathcal{L}_{\text{DW}}^{\text{W}}$ \ & mIoU (\%) \\ \midrule
Base   &               &               &               & 55.1 \\  
I  &$\checkmark$   &               &               & 55.9 \\
II   &               &$\checkmark$   & $\checkmark$  & 62.4 \\ 
III &$\checkmark$   &$\checkmark$   &               & 62.4 \\
IV  &$\checkmark$   &               & $\checkmark$  & 58.1 \\ \midrule 
V  & $\checkmark$ & $\checkmark$   & $\checkmark$  & 64.7 \\
\bottomrule
\end{tabular}
\caption{Ablation of SFC components. Base and I report the mIoU of $\mathcal{M}_{\mathbf{W}}$ and others report the mIoU of $\mathcal{M}_{\text{final}}$.}
\label{tabel_contributions}
\end{minipage} ~~
\begin{minipage}[t]{0.3\textwidth}
\makeatletter\def\@captype{table}
\centering
\setlength{\tabcolsep}{0.5mm}
\begin{tabular}{cccc}
\toprule
 &$\mathcal{L}_{\text{DW}}^{\text{P}}$ \ & $\mathcal{L}_{\text{DW}}^{\text{W}}$ \ & mIoU (\%) \\ \midrule
VI   &   &       & 62.0          \\
VII   & $\checkmark$  &               & 63.7          \\
VIII   &     & $\checkmark$ & 63.6          \\
\midrule
V   &$\checkmark$   &$\checkmark$   &64.7  \\
\midrule
\end{tabular}
\caption{Effect of $DC$. \checkmark \ indicates the presence of $DC$. VI: `Base' with plain consistency loss. mIoU of $\mathcal{M}_{\text{final}}$ is reported.}
\label{abla_ROC}
\end{minipage} ~~
\begin{minipage}[t]{0.32\textwidth}
\makeatletter\def\@captype{table}
\centering
\setlength{\tabcolsep}{0.5mm}
\small
\begin{tabular}{@{}cccc@{}}
\toprule
\multirow{2}{*}{} & \multirow{2}{*}{\begin{tabular}[c]{@{}c@{}}Classifier \\ weight\end{tabular}} \ & \multirow{2}{*}{Prototype} \ & \multirow{2}{*}{mIou (\%)}     \\
                  &                                                                               &                            &                                \\ \midrule
IX                 & $\checkmark$                                                                  &                            & 64.2                           \\
X                 &                                                                               & $\checkmark$               & 62.5                           \\
\midrule
V                 & $\checkmark$                                                                  & $\checkmark$               & 64.7 \\ \bottomrule
\end{tabular}
\caption{Ablation of inference stage. IX: Inference with classifier weight CAM $\mathcal{M}_{\mathbf{W}}$. X: Inference with prototype CAM $\mathcal{M}_{\mathbf{P}}$. V: Inference with $\mathcal{M}_{\text{final}}$.}
\label{abla_infer}
\end{minipage}
\end{table*}

\subsection{Ablation Studies}
In Table \ref{tabel_contributions}, we first verify the effectiveness of SFC components, \emph{i.e.}, Image Bank Re-sampling (IBR) and Multi-scaled Distribution-Weighted (MSDW) consistency Loss (including $\mathcal{L}_{\text{DW}}^{\text{P}}$ and $\mathcal{L}_{\text{DW}}^{\text{W}}$). `Base' is the classifier weight CAM in Eq.\eqref{pcmeq}. In Setting I, IBR increases the mIoU of `Base' CAM by 0.8\%, showing increasing the tail-class sampling frequency can alleviate the over-/under-activation issues. In Setting II, the CAM generated by SFC without IBR has a lower mIoU score than SFC (Setting V) by 2.3\%, showing increasing the tail-class sampling frequency can boost the effectiveness of $\mathcal{L}_{\text{MSDW}}$. The result of setting III shows $\mathcal{L}_{\text{DW}}^{\text{W}}$ can boost the performance improvement brought by $\mathcal{L}_{\text{DW}}^{\text{P}}$. However, setting IV indicates that using $\mathcal{L}_{\text{DW}}^{\text{W}}$ alone fails to calibrate the shared features in the down-scaled feature space and its performance drops significantly.
Table \ref{abla_ROC} studies the effectiveness of $DC$ in Eq.~\eqref{ROCeq}. Setting VI shows the SFC performance without $DC$ in both of  $\mathcal{L}_{\text{DW}}^{\text{P}}$ and $\mathcal{L}_{\text{DW}}^{\text{W}}$. Settings VII and VIII show the performances of removing $DC$ only from $\mathcal{L}_{\text{DW}}^{\text{W}}$ or $\mathcal{L}_{\text{DW}}^{\text{P}}$. The results show that $DC$ effectively adjusts the consistency loss weights of each class, bringing significant improvement compared with the plain consistency loss.
\begin{figure}[t!]
\centering
    \includegraphics[height=2.3cm]{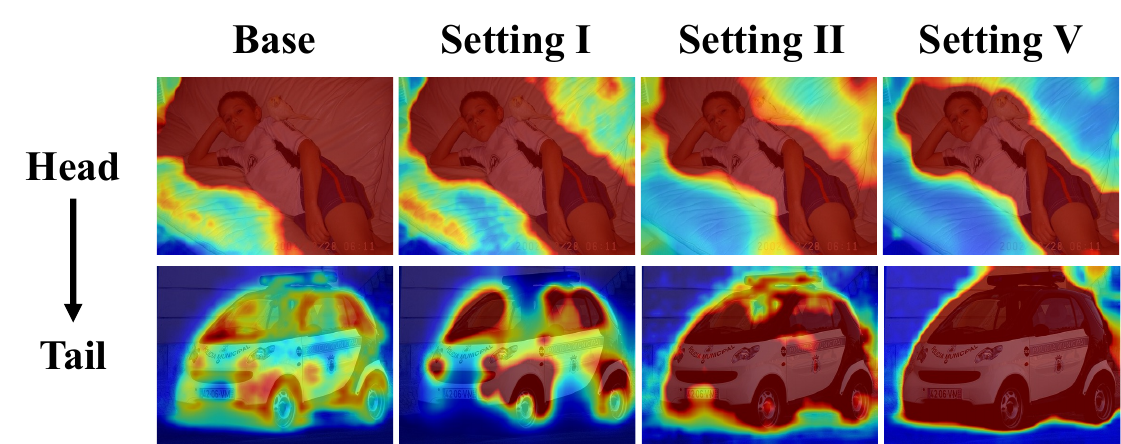}
    \caption{CAM visualization results under different settings.}
    \label{fig_analysis_2}
\end{figure}
Table \ref{abla_infer} shows that the CAM combination $\mathcal{M}_{\text{final}}$ in Eq.~\eqref{eqinfer} (setting V) has the highest performance compared with those using $\mathcal{M}_{\mathbf{W}}$ or $\mathcal{M}_{\mathbf{P}}$ alone during inference, demonstrating it is better to complement $\mathcal{M}_{\mathbf{P}}$ to $\mathcal{M}_{\mathbf{W}}$ for SFC.
\begin{table}[t!]
\centering
\small
\begin{tabular}{lccccc}
\toprule[1pt]
Class sets & Overall & Many & Medium & Few\\
\midrule
VI & 6.9 &5.6&6.5&8.6\\
V & 9.6 &9.5&6.5&12.9 \\
\bottomrule
\end{tabular}
\caption{Average mIoU gains of different class sets. The class sets definitions follow \cite{wu2020distribution}.}
\label{label_longtail}
\end{table}
Table~\ref{label_longtail} shows the average performance gains on different class sets with or without our $DC$ coefficient. We can see the plain consistency loss (setting VI) only achieves almost the same gains across `Many', `Medium', and `Few' classes. However, the head and tail classes (\emph{i.e.}, `Many' and `Few' classes) actually need higher magnitudes of consistency loss to overcome the over-/under-activation issues. With the help of $DC$ coefficient (setting V), the head and tail classes achieve more mIoU gains.

Besides, we also study the qualitative effects of SFC components in Fig.~\ref{fig_analysis_2}. It can be seen that Base with IBR (setting I) improves the CAM as it increases the tail-class sampling frequency and calibrates the shared features in classifier weights. However, the improvements are limited  (\emph{e.g.}, the shared feature `wheel' is not activated) as IBR cannot balance the training data completely. When only using $\mathcal{L}_{\text{MSDW}}$ (setting II), the CAMs are improved significantly but still not perfect, as the tail-class sampling frequencies and optimizing frequency of $\mathcal{L}_{\text{MSDW}}$ are low. By using the complete SFC (setting V), we can achieve decent CAM results.

\section{Conclusion}
In this paper, we first demonstrate that shared features can cause over-/under-activation issues in CAM generation under a long-tailed scenario and then propose a novel Shared Feature Calibration (SFC) method for solving such issues, achieving new state-of-the-art performances. Our work provides a new perspective for improving CAM accuracy in image-level weakly supervised semantic segmentation, and other possible solutions will be investigated in future work.

\section{Acknowledgements}
This work was supported by the National Key R\&D Program of China (No.2022YFE0200300), the National Natural Science Foundation of China (No. 61972323, 62331003), Suzhou Basic Research Program (SYG202316)  and XJTLU REF-22-01-010, XJTLU AI University Research Centre, Jiangsu Province Engineering Research Centre of Data Science and Cognitive Computation at XJTLU and SIP AI innovation platform (YZCXPT2022103), Suzhou Municipal Key Laboratory for Intelligent Virtual Engineering (SZS2022004). 

\bibliography{aaai24}

\begin{thebibliography}{46}
\providecommand{\natexlab}[1]{#1}

\bibitem[{Ahn, Cho, and Kwak(2019)}]{ahn2019weakly}
Ahn, J.; Cho, S.; and Kwak, S. 2019.
\newblock Weakly supervised learning of instance segmentation with inter-pixel relations.
\newblock In \emph{CVPR}.

\bibitem[{Chen et~al.(2023)Chen, Cong, Yuxuan, Ip, and Kwong}]{chen2023saving}
Chen, J.; Cong, R.; Yuxuan, L.; Ip, H.; and Kwong, S. 2023.
\newblock Saving 100x Storage: Prototype Replay for Reconstructing Training Sample Distribution in Class-Incremental Semantic Segmentation.
\newblock In \emph{NeurIPS}.

\bibitem[{Chen et~al.(2016)Chen, Papandreou, Kokkinos, Murphy, and Yuille}]{chen2017deeplab}
Chen, L.-C.; Papandreou, G.; Kokkinos, I.; Murphy, K.~P.; and Yuille, A.~L. 2016.
\newblock DeepLab: Semantic Image Segmentation with Deep Convolutional Nets, Atrous Convolution, and Fully Connected CRFs.
\newblock \emph{PAMI}, 40: 834--848.

\bibitem[{Chen et~al.(2022{\natexlab{a}})Chen, Yang, Lai, and Xie}]{chen2022self}
Chen, Q.; Yang, L.; Lai, J.-H.; and Xie, X. 2022{\natexlab{a}}.
\newblock Self-supervised image-specific prototype exploration for weakly supervised semantic segmentation.
\newblock In \emph{CVPR}.

\bibitem[{Chen et~al.(2022{\natexlab{b}})Chen, Wang, Wu, Hua, Zhang, and Sun}]{chen2022class}
Chen, Z.; Wang, T.; Wu, X.; Hua, X.-S.; Zhang, H.; and Sun, Q. 2022{\natexlab{b}}.
\newblock Class re-activation maps for weakly-supervised semantic segmentation.
\newblock In \emph{CVPR}.

\bibitem[{Deng et~al.(2009)Deng, Dong, Socher, Li, Li, and Fei-Fei}]{deng2009imagenet}
Deng, J.; Dong, W.; Socher, R.; Li, L.-J.; Li, K.; and Fei-Fei, L. 2009.
\newblock Imagenet: A large-scale hierarchical image database.
\newblock In \emph{CVPR}.

\bibitem[{Du et~al.(2022)Du, Fu, Liu, and Wang}]{du2022weakly}
Du, Y.; Fu, Z.; Liu, Q.; and Wang, Y. 2022.
\newblock Weakly supervised semantic segmentation by pixel-to-prototype contrast.
\newblock In \emph{CVPR}.

\bibitem[{Everingham et~al.(2010)Everingham, Van~Gool, Williams, Winn, and Zisserman}]{everingham2010pascal}
Everingham, M.; Van~Gool, L.; Williams, C.~K.; Winn, J.; and Zisserman, A. 2010.
\newblock The pascal visual object classes (voc) challenge.
\newblock \emph{IJCV}, 88: 303--338.

\bibitem[{Hariharan et~al.(2011)Hariharan, Arbel{\'a}ez, Bourdev, Maji, and Malik}]{hariharan2011semantic}
Hariharan, B.; Arbel{\'a}ez, P.; Bourdev, L.; Maji, S.; and Malik, J. 2011.
\newblock Semantic contours from inverse detectors.
\newblock In \emph{ICCV}.

\bibitem[{He et~al.(2016)He, Zhang, Ren, and Sun}]{he2016deep}
He, K.; Zhang, X.; Ren, S.; and Sun, J. 2016.
\newblock Deep residual learning for image recognition.
\newblock In \emph{CVPR}.

\bibitem[{Hou, Yu, and Tao(2022)}]{hou2022batchformer}
Hou, Z.; Yu, B.; and Tao, D. 2022.
\newblock Batchformer: Learning to explore sample relationships for robust representation learning.
\newblock In \emph{CVPR}.

\bibitem[{Kr{\"a}henb{\"u}hl and Koltun(2011)}]{krahenbuhl2011efficient}
Kr{\"a}henb{\"u}hl, P.; and Koltun, V. 2011.
\newblock Efficient inference in fully connected crfs with gaussian edge potentials.
\newblock In \emph{NeurIPS}.

\bibitem[{Lee et~al.(2021{\natexlab{a}})Lee, Choi, Mok, and Yoon}]{lee2021reducing}
Lee, J.; Choi, J.; Mok, J.; and Yoon, S. 2021{\natexlab{a}}.
\newblock Reducing information bottleneck for weakly supervised semantic segmentation.
\newblock In \emph{NeurIPS}.

\bibitem[{Lee, Kim, and Yoon(2021)}]{lee2021anti}
Lee, J.; Kim, E.; and Yoon, S. 2021.
\newblock Anti-adversarially manipulated attributions for weakly and semi-supervised semantic segmentation.
\newblock In \emph{CVPR}.

\bibitem[{Lee, Kim, and Shim(2022)}]{lee2022threshold}
Lee, M.; Kim, D.; and Shim, H. 2022.
\newblock Threshold matters in WSSS: manipulating the activation for the robust and accurate segmentation model against thresholds.
\newblock In \emph{CVPR}.

\bibitem[{Lee et~al.(2021{\natexlab{b}})Lee, Lee, Lee, and Shim}]{lee2021railroad}
Lee, S.; Lee, M.; Lee, J.; and Shim, H. 2021{\natexlab{b}}.
\newblock Railroad is not a train: Saliency as pseudo-pixel supervision for weakly supervised semantic segmentation.
\newblock In \emph{CVPR}.

\bibitem[{Li et~al.(2022)Li, Jie, Wang, Wei, and Ma}]{li2022expansion}
Li, J.; Jie, Z.; Wang, X.; Wei, X.; and Ma, L. 2022.
\newblock Expansion and shrinkage of localization for weakly-supervised semantic segmentation.
\newblock In \emph{NeurIPS}.

\bibitem[{Li and Monga(2020)}]{li2020group}
Li, X.; and Monga, V. 2020.
\newblock Group based deep shared feature learning for fine-grained image classification.
\newblock \emph{arXiv preprint arXiv:2004.01817}.

\bibitem[{Li et~al.(2021)Li, Zhou, Li, Zhou, and Zhang}]{li2021group}
Li, X.; Zhou, T.; Li, J.; Zhou, Y.; and Zhang, Z. 2021.
\newblock Group-wise semantic mining for weakly supervised semantic segmentation.
\newblock In \emph{AAAI}.

\bibitem[{Lin et~al.(2014)Lin, Maire, Belongie, Hays, Perona, Ramanan, Doll{\'a}r, and Zitnick}]{lin2014microsoft}
Lin, T.-Y.; Maire, M.; Belongie, S.; Hays, J.; Perona, P.; Ramanan, D.; Doll{\'a}r, P.; and Zitnick, C.~L. 2014.
\newblock Microsoft coco: Common objects in context.
\newblock In \emph{ECCV}.

\bibitem[{Lin et~al.(2023)Lin, Chen, Wang, Wu, Li, Lin, Liu, and He}]{lin2023clip}
Lin, Y.; Chen, M.; Wang, W.; Wu, B.; Li, K.; Lin, B.; Liu, H.; and He, X. 2023.
\newblock Clip is also an efficient segmenter: A text-driven approach for weakly supervised semantic segmentation.
\newblock In \emph{CVPR}.

\bibitem[{Liu et~al.(2021)Liu, Gao, Sun, Zhao, Yang, Qin, and Meng}]{liu2021infrared}
Liu, F.; Gao, C.; Sun, Y.; Zhao, Y.; Yang, F.; Qin, A.; and Meng, D. 2021.
\newblock Infrared and Visible Cross-Modal Image Retrieval Through Shared Features.
\newblock \emph{TCSVT}, 31: 4485--4496.

\bibitem[{Long, Shelhamer, and Darrell(2015)}]{long2015fully}
Long, J.; Shelhamer, E.; and Darrell, T. 2015.
\newblock Fully convolutional networks for semantic segmentation.
\newblock In \emph{CVPR}.

\bibitem[{Minaee et~al.(2021)Minaee, Boykov, Porikli, Plaza, Kehtarnavaz, and Terzopoulos}]{minaee2021image}
Minaee, S.; Boykov, Y.; Porikli, F.; Plaza, A.; Kehtarnavaz, N.; and Terzopoulos, D. 2021.
\newblock Image segmentation using deep learning: A survey.
\newblock \emph{PAMI}, 44(7): 3523--3542.

\bibitem[{Peng, He, and Zhao(2017)}]{peng2017object}
Peng, Y.; He, X.; and Zhao, J. 2017.
\newblock Object-part attention model for fine-grained image classification.
\newblock \emph{TIP}, 27(3): 1487--1500.

\bibitem[{Sun et~al.(2020)Sun, Wang, Dai, and Van~Gool}]{sun2020mining}
Sun, G.; Wang, W.; Dai, J.; and Van~Gool, L. 2020.
\newblock Mining cross-image semantics for weakly supervised semantic segmentation.
\newblock In \emph{ECCV}.

\bibitem[{Tan et~al.(2021)Tan, Lu, Zhang, Yin, and Li}]{tan2021equalization}
Tan, J.; Lu, X.; Zhang, G.; Yin, C.; and Li, Q. 2021.
\newblock Equalization loss v2: A new gradient balance approach for long-tailed object detection.
\newblock In \emph{CVPR}.

\bibitem[{Vasconcelos, Birodkar, and Dumoulin(2022)}]{vasconcelos2022proper}
Vasconcelos, C.; Birodkar, V.; and Dumoulin, V. 2022.
\newblock Proper reuse of image classification features improves object detection.
\newblock In \emph{CVPR}, 13628--13637.

\bibitem[{Wang et~al.(2020)Wang, Zhang, Kan, Shan, and Chen}]{wang2020self}
Wang, Y.; Zhang, J.; Kan, M.; Shan, S.; and Chen, X. 2020.
\newblock Self-supervised equivariant attention mechanism for weakly supervised semantic segmentation.
\newblock In \emph{CVPR}.

\bibitem[{Wu et~al.(2020)Wu, Huang, Liu, Wang, and Lin}]{wu2020distribution}
Wu, T.; Huang, Q.; Liu, Z.; Wang, Y.; and Lin, D. 2020.
\newblock Distribution-balanced loss for multi-label classification in long-tailed datasets.
\newblock In \emph{ECCV}.

\bibitem[{Xie et~al.(2022)Xie, Hou, Ye, and Shen}]{xie2022clims}
Xie, J.; Hou, X.; Ye, K.; and Shen, L. 2022.
\newblock CLIMS: cross language image matching for weakly supervised semantic segmentation.
\newblock In \emph{CVPR}.

\bibitem[{Xu et~al.(2021)Xu, Ouyang, Bennamoun, Boussaid, Sohel, and Xu}]{xu2021leveraging}
Xu, L.; Ouyang, W.; Bennamoun, M.; Boussaid, F.; Sohel, F.; and Xu, D. 2021.
\newblock Leveraging auxiliary tasks with affinity learning for weakly supervised semantic segmentation.
\newblock In \emph{CVPR}.

\bibitem[{Xu et~al.(2022)Xu, Ouyang, Bennamoun, Boussaid, and Xu}]{xu2022multi}
Xu, L.; Ouyang, W.; Bennamoun, M.; Boussaid, F.; and Xu, D. 2022.
\newblock Multi-class token transformer for weakly supervised semantic segmentation.
\newblock In \emph{CVPR}.

\bibitem[{Xu et~al.(2023)Xu, Wang, Sun, Xu, Meng, and Zhang}]{xu2023self}
Xu, R.; Wang, C.; Sun, J.; Xu, S.; Meng, W.; and Zhang, X. 2023.
\newblock Self correspondence distillation for end-to-end weakly-supervised semantic segmentation.
\newblock In \emph{AAAI}.

\bibitem[{Yao et~al.(2017)Yao, Zhang, Yan, Zhang, Li, and Tian}]{yao2017autobd}
Yao, H.; Zhang, S.; Yan, C.; Zhang, Y.; Li, J.; and Tian, Q. 2017.
\newblock AutoBD: Automated bi-level description for scalable fine-grained visual categorization.
\newblock \emph{TIP}, 27(1): 10--23.

\bibitem[{Yao et~al.(2021)Yao, Chen, Xie, Zhang, Shen, Wu, Tang, and Zhang}]{yao2021non}
Yao, Y.; Chen, T.; Xie, G.-S.; Zhang, C.; Shen, F.; Wu, Q.; Tang, Z.; and Zhang, J. 2021.
\newblock Non-salient region object mining for weakly supervised semantic segmentation.
\newblock In \emph{CVPR}.

\bibitem[{Yoon et~al.(2022)Yoon, Kweon, Cho, Kim, and Yoon}]{yoon2022adversarial}
Yoon, S.-H.; Kweon, H.; Cho, J.; Kim, S.; and Yoon, K.-J. 2022.
\newblock Adversarial erasing framework via triplet with gated pyramid pooling layer for weakly supervised semantic segmentation.
\newblock In \emph{ECCV}.

\bibitem[{Zhang et~al.(2021{\natexlab{a}})Zhang, Xiao, Jiao, Wei, and Zhao}]{zhang2021affinity}
Zhang, B.; Xiao, J.; Jiao, J.; Wei, Y.; and Zhao, Y. 2021{\natexlab{a}}.
\newblock Affinity attention graph neural network for weakly supervised semantic segmentation.
\newblock \emph{PAMI}, 44(11): 8082--8096.

\bibitem[{Zhang et~al.(2023{\natexlab{a}})Zhang, Xiao, Wei, and Zhao}]{zhang2023credible}
Zhang, B.; Xiao, J.; Wei, Y.; and Zhao, Y. 2023{\natexlab{a}}.
\newblock Credible Dual-Expert Learning for Weakly Supervised Semantic Segmentation.
\newblock \emph{IJCV}, 131: 1892 -- 1908.

\bibitem[{Zhang et~al.(2020)Zhang, Zhang, Tang, Hua, and Sun}]{zhang2020causal}
Zhang, D.; Zhang, H.; Tang, J.; Hua, X.-S.; and Sun, Q. 2020.
\newblock Causal intervention for weakly-supervised semantic segmentation.
\newblock In \emph{NeurIPS}.

\bibitem[{Zhang et~al.(2021{\natexlab{b}})Zhang, Gu, Zhang, and Dai}]{zhang2021complementary}
Zhang, F.; Gu, C.; Zhang, C.; and Dai, Y. 2021{\natexlab{b}}.
\newblock Complementary patch for weakly supervised semantic segmentation.
\newblock In \emph{ICCV}.

\bibitem[{Zhang et~al.(2023{\natexlab{b}})Zhang, Wang, Kang, Chen, and Wei}]{zhang2023slca}
Zhang, G.; Wang, L.; Kang, G.; Chen, L.; and Wei, Y. 2023{\natexlab{b}}.
\newblock SLCA: Slow Learner with Classifier Alignment for Continual Learning on a Pre-trained Model.
\newblock \emph{arXiv preprint arXiv:2303.05118}.

\bibitem[{Zhang et~al.(2022)Zhang, Gao, Fang, Jiao, and Wei}]{zhang2022mining}
Zhang, Z.; Gao, G.; Fang, Z.; Jiao, J.; and Wei, Y. 2022.
\newblock Mining Unseen Classes via Regional Objectness: A Simple Baseline for Incremental Segmentation.
\newblock \emph{NeurIPS}, 35: 24340--24353.

\bibitem[{Zheng et~al.(2017)Zheng, Fu, Mei, and Luo}]{zheng2017learning}
Zheng, H.; Fu, J.; Mei, T.; and Luo, J. 2017.
\newblock Learning multi-attention convolutional neural network for fine-grained image recognition.
\newblock In \emph{ICCV}.

\bibitem[{Zhou et~al.(2016)Zhou, Khosla, Lapedriza, Oliva, and Torralba}]{zhou2016learning}
Zhou, B.; Khosla, A.; Lapedriza, A.; Oliva, A.; and Torralba, A. 2016.
\newblock Learning deep features for discriminative localization.
\newblock In \emph{CVPR}.

\bibitem[{Zhu et~al.(2023)Zhu, Wei, Liang, Zhang, and Zhao}]{zhu2023ctp}
Zhu, H.; Wei, Y.; Liang, X.; Zhang, C.; and Zhao, Y. 2023.
\newblock CTP: Towards Vision-Language Continual Pretraining via Compatible Momentum Contrast and Topology Preservation.
\newblock In \emph{ICCV}, 22257--22267.

\end{thebibliography}

\end{document}